# Automatic Detection of Arousals during Sleep using Multiple Physiological Signals

Saman Parvaneh[1], Jonathan Rubin[1], Ali Samadani[1], Gajendra Katuwal[1]

[1]Philips Research North America, Cambridge, MA, USA

## Abstract

*The visual scoring of arousals during sleep routinely conducted by sleep experts is a challenging task warranting an automatic approach. This paper presents an algorithm for automatic detection of arousals during sleep. Using the Physionet/CinC Challenge dataset, an 80-20% subject-level split was performed to create in-house training and test sets, respectively. The data for each subject in the training set was split to 30-second epochs with no overlap. A total of 428 features from EEG, EMG, EOG, airflow, and SaO2 in each epoch were extracted and used for creating subject-specific models based on an ensemble of bagged classification trees, resulting in 943 models. For marking arousal and non-arousal regions in the test set, the data in the test set was split to 30-second epochs with 50% overlaps. The average of arousal probabilities from different patient-specific models was assigned to each 30-second epoch and then a sample-wise probability vector with the same length as test data was created for model evaluation. Using the PhysioNet/CinC Challenge 2018 scoring criteria, AUPRCs of 0.25 and 0.21 were achieved for the in-house test and blind test sets, respectively.*

## 1. Introduction

Arousals during sleep can cause awakening or sleep stage shifts [1]. Arousals are naturally occurring micro events [2], but they can become pathological when the frequency of occurrence increases beyond the normal limit [1, 2]. Arousals are found to be associated with the pathophysiology of several sleep disorders [2].

The visual scoring of arousals routinely conducted by sleep experts is a time consuming and cumbersome task warranting an automatic approach. The main goal of the 2018 PhysioNet/CinC Challenge was to use available vital signs including airflow, electroencephalography (EEG), electromyography (EMG), electrocardiology (ECG), and oxygen saturation (SaO2) to correctly classify target arousal regions. This paper presents an algorithm for automatic detection of arousal regions during sleep using ensemble of subject-specific models created by multiple physiological signals.

## 2. Material and Method

A block diagram of our proposed method is shown in Figure 1.

### 2.1. Data

The training and test sets for the challenge include physiological signals (six channels of EEG, EOG, three channels of EMG, ECG, and SaO2) for 994 and 989 subjects, respectively. All signals except SaO2 were sampled at 200 Hz and SaO2 was resampled to 200 Hz. Non-arousal, arousal, and undefined annotations were provided in a sample-wise vector based on annotations from certified sleep technologists. Areas with undefined annotations will not be scored. Details about the challenge dataset can be found in a paper written by Ghasemi et al. [3].

The training set was split to 796 (80%) and 198 (20%) subjects as in-house train and test subjects. The in-house train-set was used for model training and validation and in-house test-set was used for assessing algorithm performance independent from the blind challenge test set.

### 2.2. Subject-specific Models

A separate model was created for each subject in the training set (subject-specific modelling). To this end, subjects whose signal included both arousal and non-arousal annotations were used for subject-specific modelling. To train subject-specific models, a 30 seconds wide moving window with no overlap was used to create 30 seconds epochs required for training a model. For epochs with multiple annotation classes (arousal, non-arousal and/or undefined), the most frequent annotation was used as the label for the epoch. For example, if 70% and 30% of an epoch had arousal and undefined annotations, the epoch was annotated as arousal. Furthermore, an epoch was excluded from training, if the entire epoch was annotated as undefined.

In the training phase, in-house training set was used for training models and result on in-house test set was used for selecting optimum physiological signals, evaluating extracted features, and selecting the best classifier. After having an acceptable trained model, additional models were created on the in-house test set and combined with models on in-house train set for final evaluation on blind challenge test set.

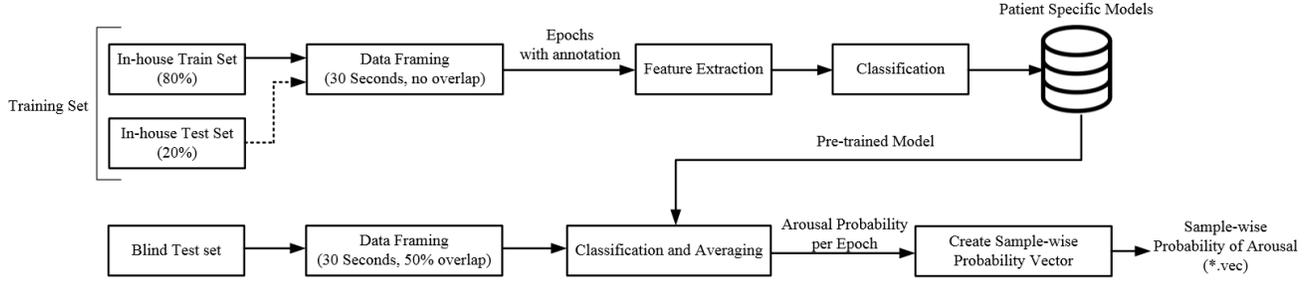

Figure 1. Block diagram of the proposed algorithm.

### 2.2.1. Feature Engineering

A total of 428 features were extracted from EEG, EMG, EOG, Airflow, and SaO2.

- *EEG (119 features from all six channels)*: frequency spectral power in different bands (Delta: 1-3Hz, Theta: 4-7Hz, Alpha1: 8-9Hz, Alpha2: 10-12Hz, Beta1: 13-17Hz, Beta2: 18-30Hz, Gamma1: 31-40Hz, Gamma2: 41-50Hz, higher bands1: 51-70Hz, and higher bands2: 71-100Hz), mean of power in different bands across six channels, correlation between channels in the temporal and frequency domains were extracted from six EEG channels [4].
- *SaO2 (9 features)*: mean, standard deviation, coefficient of variation (CV), skewness, and kurtosis of SaO2 were calculated. Additionally, percentage of time that SaO2≥96, 90≤SaO2<96, 80≤SaO2<90, and SaO2<80 in 30 seconds windows were used as SaO2 features to quantify duration of normal, mild, moderate, and severe oxygen desaturation (i.e. reduction in blood oxygen level), respectively.
- *EMG (64 features for each EMG channels)*: statistical measures (i.e., average, minimum, maximum, range, variance, CV, skewness, and kurtosis), integral of absolute value (IAV), mean of absolute value (MAV), zero crossing rate (ZCR), slope sign changes (SSC), waveform length (WL), root mean square (RMS), average rectified value (ARV), Willison amplitude (WAMP) [5-7] as well as time-domain power spectral moments were extracted for each EMG channels [8]. Number of EMG data points in each bin of a histogram with 20 bins and their relative frequency (number of data points in each bin normalized to the total number of points) were also considered as features.
- *EOG (15 features)*: After smoothing EOG, statistical measures (i.e. minimum, maximum, range, average, median, skewness, kurtosis), IAV, energy, RMS, form factor of signal, ratio of standard deviation of first derivative of EOG to standard deviation of EOG, ratio of standard deviation of second derivative of EOG to standard deviation of EOG, integral of absolute value of derivative were calculated [9].
- *Airflow (68 features)*: smoothness of airflow waveform was quantified by standard deviation and CV of the differences between adjacent samples of airflow waveform as well as lag one autocorrelation. Difference between areas under positive and negative airflow is another feature extracted from airflow waveform. Furthermore, the same features used with EMG were also used with smoothed airflow waveform.
- *Interaction features (25 features)*: Cross correlation between every pair of SaO2, smoothed airflow, chest EMG, abdominal EMG, and chin EMG were calculated and lags corresponding to the maximum absolute value of cross correlations were extracted as measures that quantify interaction between physiological signals (10 features). A similar approach was used for six EEG channels, leading to 15 features.

In addition to the above-mentioned features, heart rate and heart rate variability (HRV) features reported in previous research studies were extracted from ECG [10-15]. Since the contribution of proposed features (e.g. SDNN, RMSSD, heart rate asymmetry, featured from Triangle Phase Space Mapping, features from Parabolic Phase Space Mapping) to the performance of algorithm on in-house test set was not significant, it was decided to exclude them from feature extraction to reduce running time for training and evaluating models.

### 2.2.2. Classifier and Model Training

An ensemble of 30 bagged classification trees was determined as the best classifier based on performance on in-house test set and was used for classification of arousal versus non-arousal epochs. As mentioned above, one model was trained for each subject in the training set and the trained model was saved in the patient specific model database (total of 943 models).

### 2.2.3. Algorithm Evaluation

For algorithm evaluation, a 30 seconds wide moving window and 50 percent overlap was used in the in-house and blind test sets. After feature extraction in each epoch, the resulting feature vector was processed with all the subject-specific models in the model database and the average of classification probabilities was assigned to all samples in the epoch under study. At the end, the epoch-specific arousal probability was used to create sample-wise portability of arousal (*.vec file for model evaluation).

Of note, only models created on the in-house train set were used on the in-house test set for initial algorithm evaluation. After finding the optimum classification configuration (physiological signals, features, and classifier) in the initial algorithm evaluation, models created using the entire training set (both in-house training and test sets) were used for evaluation of the blind test set.

The performance of the algorithm in the binary classification of target arousal and non-arousal regions was measured in terms of the gross area under the precession-recall curve (AUPRC). Additionally, the gross area under the receiver operating characteristic curve (AUROC) was reported. For independent evaluation of algorithm from blind test set, in-house test set was used. For the final scoring during official phase of the challenge, the performance was evaluated on a random subset of blind hidden and the whole blind test set.

### 3. Results and Discussion

The performance of the proposed method on the in-house test set and hidden test set is reported in Table 1. The best result achieved by the proposed algorithm at the official phase of the challenge on the in-house test set and on the challenge blind test dataset were AUPRC of 0.255 and 0.21, respectively.

Table 1. Algorithm performance on the in-house test sets and blind test set. AUROC and AUPRC are the gross area under ROC and precision-recall curves, respectively.

| Dataset | AUROC | AUPRC |
|---|---|---|
| In-house test set | 0.826 | 0.255 |
| Blind test set | 0.794 | 0.21 |

In the following paragraphs, some of the findings of other approaches investigated during the CinC challenge are discussed.

- Adding heart rate and HRV features to the subject-specific models did not lead to a significant improvement in algorithm performance in the in-house test set. Therefore, they were removed from the final algorithm to reduce running time for extracting features required for QRS detection and ECG signal quality check.
- A subject-independent model for classification of arousal versus non-arousal epochs (one model that separates these two groups) was created by using the above-mentioned features. Although the performance of developed model was promising in the classification task, the performance for separating arousal and non-arousal regions was lower than the subject-specific model. Further exploration in future studies is required to improve performance of subject-independent model for marking arousal and non-arousal regions (segmentation).
- Using the sleep stages as input features was considered in a previous study [2]. We initially used the sleep annotation in the training set as input features to the classifier. However, the inclusion of sleep stages was not considered for our final entry as creating a sleep staging model was required due to the lack of sleep annotations for the test set. In the future, the impact of adding sleep stages in classification of arousal and non-arousal regions will be explored using sleep staging models.

### 4. Conclusion

In this study, ensemble of subject-specific models was tested for automatic detection of arousals regions during sleep using multiple physiological signals. The performance of the proposed algorithm and its simplicity encourages us to further improve the algorithm in the future using additional features as well as by creating and augmenting new deep neural network models for different physiological signals.

### References

[1] F. De Carli, L. Nobili, P. Gelcich, and F. Ferrillo, "A method for the automatic detection of arousals during sleep," *Sleep,* vol. 22, pp. 561-572, 1999.
[2] M. Olsen, L. D. Schneider, J. Cheung, P. E. Peppard, P. J. Jennum, E. Mignot*, et al.*, "Automatic, electrocardiographic-based detection of autonomic arousals and their association with cortical arousals, leg movements, and respiratory events in sleep," *Sleep,* vol. 41, p. zsy006, 2018.
[3] M. M. Ghasemi, E. M. Benjamin, L.-w. H. Lehman, C. Song, Q. Li, H. Sun*, et al.*, "You Snooze, You Win: the PhysioNet/Computing in Cardiology Challenge 2018," presented at the Computing in Cardiology, Maastricht, Netherlands, 2018.
[4] G. Jones. (2016). *EEG Seizure Prediction*. Available: https://github.com/garethjns/Kaggle-EEG
[5] A. Phinyomark, C. Limsakul, and P. Phukpattaranont, "A novel feature extraction for robust EMG pattern recognition," *arXiv preprint arXiv:0912.3973,* 2009.
[6] N. Nazmi, M. A. Abdul Rahman, S.-I. Yamamoto, S. A. Ahmad, H. Zamzuri, and S. A. Mazlan, "A review of

classification techniques of EMG signals during isotonic and isometric contractions," *Sensors,* vol. 16, p. 1304, 2016.

[7] M. Naji, M. Firoozabadi, and S. Kahrizi, "Evaluation of EMG features of trunk muscles during flexed postures," in *Biomedical Engineering (ICBME), 2012 19th Iranian Conference of*, 2012, pp. 71-74.

[8] A. H. Al-Timemy, R. N. Khushaba, G. Bugmann, and J. Escudero, "Improving the performance against force variation of EMG controlled multifunctional upper-limb prostheses for transradial amputees," *IEEE Trans. Neural Syst. Rehabil. Eng,* vol. 24, pp. 650-661, 2016.

[9] A. Coskun, S. Ozsen, S. Yucelbas, C. Yucelbas, G. Tezel, S. Kuccukturk*, et al.*, "Detection of REM in Sleep EOG Signals," *Indian Journal of Science and Technology,* vol. 9, 2016.

[10] S. Parvaneh, N. Toosizadeh, and S. Moharreri, "Impact of mental stress on heart rate asymmetry," in *Computing in Cardiology Conference (CinC), 2015*, 2015, pp. 793-796.

[11] S. Moharreri, S. Rezaei, N. J. Dabanloo, and S. Parvaneh, "Extended Parabolic Phase Space Mapping (EPPSM): Novel Quadratic Function for Representation of Heart Rate Variability Signal," in *Computing in Cardiology*, 2014.

[12] S. Moharreri, N. Jafarnia Dabanloo, S. Parvaneh, and A. M. Nasrabadi, "How to Interpret Psychology from Heart Rate Variability?," in *Middle East Conference on Biomedical Engineering (MECBME)*, Sharjah, UAE, 2011.

[13] S. Rezaei, S. Moharreri, S. Ghiasi, and S. Parvaneh, "Diagnosis of Sleep Apnea by Evaluating Points Distribution in Poincare Plot of RR Intervals," *Computing,* vol. 44, p. 1, 2017.

[14] S. Parvaneh, J. Rubin, A. Rahman, B. Conroy, and S. Babaeizadeh, "Analyzing single-lead short ECG recordings using dense convolutional neural networks and feature-based post-processing to detect atrial fibrillation," *Physiological measurement,* vol. 39, p. 084003, 2018.

[15] S. Moharreri, N. J. Dabanloo, S. Parvaneh, and A. M. Nasrabadi, "The Relation between Colors, Emotions and Heart using Triangle Phase Space Mapping (TPSM)," *Computers in Cardiology,* vol. 38, 2011.

Address for correspondence.
Saman Parvaneh
2 Canal Park, 3rd floor, Cambridge, MA 02141.
saman.parvaneh@philips.com